\documentclass[letterpaper, 10 pt, conference]{ieeeconf}
\usepackage[utf8]{inputenc}
\pdfoutput=1
\IEEEoverridecommandlockouts
\title{\LARGE \bf Learned Lifted Linearization Applied to Unstable Dynamic Systems Enabled by Koopman Direct Encoding}
\author{Jerry Ng$^{1}$, H. Harry Asada$^{2}$,$~\IEEEmembership{Fellow,~IEEE}$%
\thanks{This material is based upon work supported by National Science Foundation Grant NSF-CMMI 2021625.} 
\thanks{$^{1}$Jerry Ng is with Department of Mechanical Engineering at the Massachusetts Institute of Technology
        {\tt\small jerryng@mit.edu}}%
\thanks{$^{2}$H. Harry Asada is with Department of Mechanical Engineering at the Massachusetts Institute of Technology
        {\tt\small asada@mit.edu}}%
}
\usepackage[ruled]{algorithm2e}
\usepackage[font=small]{caption}

\usepackage{cite}
\usepackage{amsmath,amssymb,amsfonts}
\usepackage{algorithmic}
\usepackage{graphicx}
\usepackage{textcomp}
\usepackage{xcolor}
\usepackage{color,soul}
\usepackage{mathtools} 
\usepackage{subcaption}
\usepackage{url}
\usepackage{bondgraphs}
\usepackage{standalone}

\begin{document}

\maketitle
\begin{abstract}
This paper presents a Koopman lifting linearization method that is applicable to nonlinear dynamical systems having both stable and unstable regions. It is known that Dynamic Mode Decomposition (DMD) and its extended methods are often unable to model unstable systems accurately and reliably. Here we solve the problem through merging three methodologies: decomposition of a lifted linear system into stable and unstable modes, deep learning of a dictionary of observable functions in the separated subspaces, and a new formula for obtaining the Koopman operator, called Direct Encoding. Two sets of effective observable functions are obtained through neural net training where the training data are separated into stable and unstable trajectories. The resultant learned observables are used for lifting the state space, and a linear state transition matrix is constructed using Direct Encoding where inner products of the learned observables are computed. The proposed method shows a dramatic improvement over existing DMD and data-driven methods. Furthermore, a method is developed for determining the boundaries between stable and unstable regions.

\end{abstract}
\section{Introduction}
\label{intro}


Lifting linearization methods, such as those based on the Koopman Operator, have been used to transform nonlinear systems to linear models. The theory states that the linear model becomes exact in modeling a nonlinear system as the order of the linear model approaches infinity, though some nonlinear systems have finite order representations in lifted space \cite{mezic2017spectral, brunton2021tutorial}. The Koopman Operator models are generally constructed through data-driven methods such as Extended Dynamic Mode Decomposition (EDMD) \cite{williams2015data}. These Koopman-DMD methods have been applied to non-autonomous systems to construct linear dynamic models that allow us to apply various linear control methods, such as  linear model predictive control (MPC), to nonlinear control systems\cite{KoopmanMeetsMPC}. 

A key component necessary to constructing a Koopman Operator-based linear model is selection of the observable functions that lift the state space. Prior work has studied the use of various function families as observables, such as polynomial basis functions, radial basis functions and time delays \cite{servadio2021orthpoly, kamb2018timedelay, mauroy2019koopman} . There have also been formulas created for algorithmically determining useful observables based on the dataset \cite{manjunathobs2021, das2018liegroup}. Modern machine learning techniques have been applied to learn observable functions with significant success\cite{lusch2018deep, han2020deep, yeung2017deep}. 

However, it can be difficult to formulate accurate approximations of the Koopman Operator for nonlinear systems that produce both stable and unstable trajectories. A passive dynamic walker, for example, is essentially an unstable system, but it can walk stably if it starts within a stable region \cite{mcgeer1990}. When concerned with only the stable regions of a nonlinear system, methods have been developed to construct stable Koopman models from unstable data-driven models for systems that are known to be stable \cite{fanstabkoop2021, mamakoukas2020learning, bevanda2022diffeomorphically, han2021desko}. 
However, these methods are not applicable when needing to predict stable trajectories for a system with unstable regions. A key difficulty is to capture a proper dataset that represents diverse behaviors involved in stable and unstable trajectories. Because of the nature of unstable trajectories, a bias towards the unstable modes can often occur when creating the Koopman model.

Prior work discusses the potential for Koopman Operator models to describe unstable subspaces \cite{mezic2017spectral}. This paper presents a methodology for constructing an accurate Koopman model for nonlinear systems with both stable and unstable regions through separation of a lifted space into stable and unstable subspaces. Two sets of effective observables are learned separately and superposed to construct a complete model.
In addition, a method will be developed for determining a boundary between stable and unstable regions in the original state space by analyzing the Koopman Operator Model using modal decomposition, which is made possible with the effective construction of observable functions.

The current work presents two significant contributions. One is a novel training method of subspace specific observable generation (SSOG) via a neural network. The other is application of a new formula, called Direct Encoding \cite{asada2022}, for obtaining Koopman Operator models through inner product computations instead of least squares estimate.

In the following, the Koopman Direct Encoding is briefly described in Section \ref{preliminaries}, followed by the development of the SSOG algorithm based on space separation, observable training, and model construction using the Direct Encoding  (Section \ref{contribution}). Numerical experiments are presented in Section \ref{results}, and the results will be discussed in Section \ref{discussion}.  

\section{Koopman Direct Encoding of Nonlinear Dynamics}
\label{preliminaries}

In this section, we give a brief overview of the Koopman Operator and introduce the direct encoding method for obtaining a Koopman operator directly from nonlinear dynamics.


Consider a discrete-time dynamical system, given by

\begin{equation}
    x_{k+1} = f(x_k) 
    \label{eq:gen_nonlinear}
\end{equation}
where $x \in \mathbb{R}^n$ is the independent state variable vector representing the system, 
$f$ is a nonlinear function $f : \mathbb{R}^n \rightarrow \mathbb{R}^n$,  and $k$ is the current time step. Also consider a real-valued observable function of the state variables $g : \mathbb{R}^n \rightarrow \mathbb{R}$. The Koopman Operator is an infinite-dimensional linear operator acting on the observable function $g$ :
\begin{equation}
    (\mathcal{K} g) (x) = g(f(x)) = (g \circ f)(x)
    \label{eq: Koopman}
\end{equation}
where the Koopman operator $\mathcal{K}$ is linear, even though the dynamic system is nonlinear.




Although this Koopman operator can be defined for an observable involved in a general Banach space \cite{mauroy2020introduction}, we assume that the observable function $g$ exists in a Hilbert space on $X \subset \mathcal{R}^n$, 
\begin{equation}
    g \in \mathcal{H}
\end{equation}
Then, it can be shown that the composition $g \circ f$ in (\ref{eq: Koopman}) can be expressed with an integral kernel as


\begin{equation}
(g \circ f)(x) = \int_X \kappa(x, \xi) g(\xi) d\xi
\label{eq: Integral kernel}
\end{equation}
where $\kappa: X \times X \rightarrow C$ is a kernel that can be written by using a set of orthonormal basis functions $[\phi_1, \phi_2, \phi_3, ...]$ that span $\mathcal{H}$.
\begin{equation}
    \kappa(x, \xi) = \sum_{i=1}^\infty \phi_i[f(x)] \bar{\phi}_i(\xi)
\end{equation}
This demonstrates that the composition function $g \circ f$ is given by the linear transformation of the observable function $g \in \mathcal{H}$ \cite{asada2022}.

Let $[g_1, g_2, \cdots ]$ be an independent set of observables that spans the Hilbert space $\mathcal{H}$. Let us further assume that the compositions of $g_i$ with $f$ are involved in the Hilbert space.

\begin{equation}
    g_i \circ f \in \mathcal{H} \quad\forall i
\end{equation}
Applying the above linear transformation of the observable function (\ref{eq: Integral kernel}) to all the observables, it can be shown that a time-evolution of the observables is given as a linear transformation with an infinite-dimensional matrix $A$. 

\begin{equation}
    z[f(x)] = A z(x)
    \label{eq: z[f(x)]}
\end{equation}
where 
\begin{equation}
    z(x) = \begin{bmatrix} g_1(x) & g_2(x) & \dots\end{bmatrix}^T
\end{equation}
The matrix {$A$} is a state transition matrix that maps the lifted state from one time step to the next.

In the Direct Encoding method \cite{asada2022}, the matrix $A$ is determined by taking inner products of the observables and their compositions with the nonlinear function $f$. Post-multiplying $z^T(x)$ to both sides of (\ref{eq: z[f(x)]}) and taking integral over $X$ yields

\begin{equation}
    Q=AR
\end{equation}
where
\begin{align}
    Q &= \begin{bmatrix} \langle g_1\circ f, g_1 \rangle & \langle g_1 \circ f, g_2  & \dots \\ \langle g_2 \circ f, g_1 \rangle& \langle g_2 \circ f, g_2 \rangle & \dots \\ \vdots & \vdots & \ddots \end{bmatrix}\\
    R &= \begin{bmatrix} \langle g_1, g_1 \rangle& \langle g_1, g_2 \rangle & \dots \\ \langle g_2, g_1 \rangle & \langle g_2, g_2 \rangle & \dots \\ \vdots & \vdots & \ddots \end{bmatrix}
\end{align}
Since $[g_1, g_2, \cdots]$ are independent, the matrix $R$ is non-singular. Therefore, the $A$ matrix is given by
\begin{equation}
    A = QR^{-1}
\end{equation}

This Direct Encoding method allows us to obtain a linear model directly from a given nonlinear state equation of function $f$ and observable functions. The model is valid globally.

Although the original Koopman Operator is infinite dimensional, effective methods have been established for approximating the operator \cite{korda2020optimal,johnson2018logistic,manjunath2021universal}.





\section{Subspace Specific Observable Generation}
\label{contribution}

This section presents a novel algorithm for obtaining an accurate Koopman operator model for nonlinear systems having both stable and unstable regions. The algorithm is built upon three theoretical and technical foundations.

First, it employs the Direct Encoding formula. In DMD, including its variants such as EDMD, the linear state transition matrix $A$ in eq. (\ref{eq: z[f(x)]}) is assumed to exist and is determined from data based on a Least Squares Estimation. This may cause a biased estimate, as addressed previously. In the Direct Encoding formula, however, the $A$ matrix is determined from the inner products of observable functions and their composition with the nonlinear state function $f$ involved in the two matrices $R$ and $Q$. Because both $g_i$ and $g_i \circ f$ are in a Hilbert space, all the inner products are guaranteed to exist and the resultant $A$ matrix provides an exact linearization that does not depend on data. The linear model is globally valid. This Direct Encoding formula is used as a foundational framework in the new algorithm.

Second, the algorithm exploits a basic property of a linear dynamical system.
If a Koopman Operator model can be constructed for a nonlinear system, then the system can be represented by
\begin{equation}
    z_{k+1} = A z_k
\end{equation}
from eqs. (\ref{eq:gen_nonlinear}) and (\ref{eq: z[f(x)]}). This linear system can be separated into its individual modes using eigendecomposition.
\begin{equation}
    z_{k+1} = V_uD_u^tW_u^T z_k + V_mD_m^t W_m^T z_k + V_sD_s^tW_s^T z_k
\end{equation}
where $V, D,$ and $W$ are the eigendecomposition of $A$, and the subscripts $u, m,$ and $s$ represent unstable, marginally stable, and stable subspaces. This decomposition in the lifted space motivates us to construct observable functions that represent the individual subspaces.


\begin{figure} [ht]
    \centering
    \includegraphics[width = 0.9\linewidth]{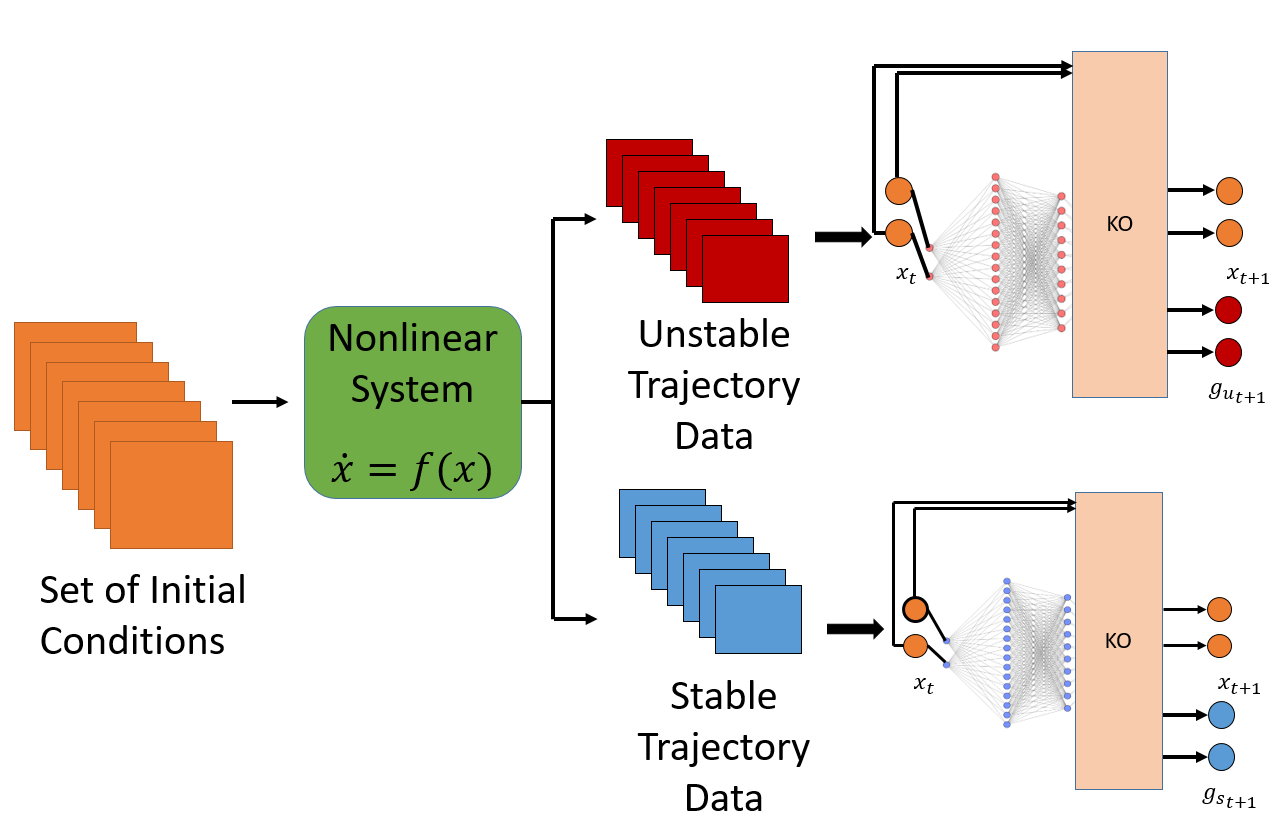}
    \caption{The training scheme utilized to generate the architecture in Fig. \ref{fig:hybrid_arch}.}
    \label{fig:training}
\end{figure}

Third, the algorithm uses neural networks to find an effective set of observables for each of the stable and unstable subspaces.
The training method, summarized in Fig \ref{fig:training}, assumes that while the underlying system has unknown characteristics, the trajectory data obtained from the system can be separated into two categories: stable/marginally stable, and unstable. Specifically, an aggregate dataset is comprised of initial states $X_k$ and the states at one time step ahead $X_{k+1}$. This can then be organized into
\begin{equation}
    X = \begin{bmatrix} X_u & X_s\end{bmatrix}
\end{equation}
for both $X_k$ and $X_{k+1}$. These subsets of data are used to train corresponding neural networks $G_u(x_i; \theta_i)$ and $G_s(x_i; \theta_i)$ and the weights of linear output layers $A_s$ and $A_u$. The function of the neural networks is to produce observables of a given state, that is
\begin{equation}
    G_*(x) = g_*(x)
\end{equation}
and 
\begin{equation}
    z_{*_{k+1}} = A_* z_{*_k}
\end{equation}
where $*$ can be replaced with $s$ or $u$, representing the specific subset of data, and $z$ is the lifted state represented by
\begin{equation}
    z_* = \begin{bmatrix} x_* & g_*(x_*) \end{bmatrix} ^T
\end{equation}

The loss function utilized in training all models is
\begin{equation}
    L(z, \hat{z}) = \frac{1}{N} \sum_{i=0}^N (z - \hat{z})^2
\end{equation}

After the training of these networks is completed with the trajectory dataset, the weights for the neural networks $G_u$ and $G_s$ no longer are updated. We then utilize the Direct Encoding method to produce a new estimation for the output layer. This new regression is computed by creating a new lifted state through concatenating the observable functions of each network with the state vector
\begin{equation}
    z = \begin{bmatrix} x & g_u & g_s \end{bmatrix}^T
\end{equation}

\begin{figure}
    \centering
    \includegraphics[width = 0.8\linewidth]{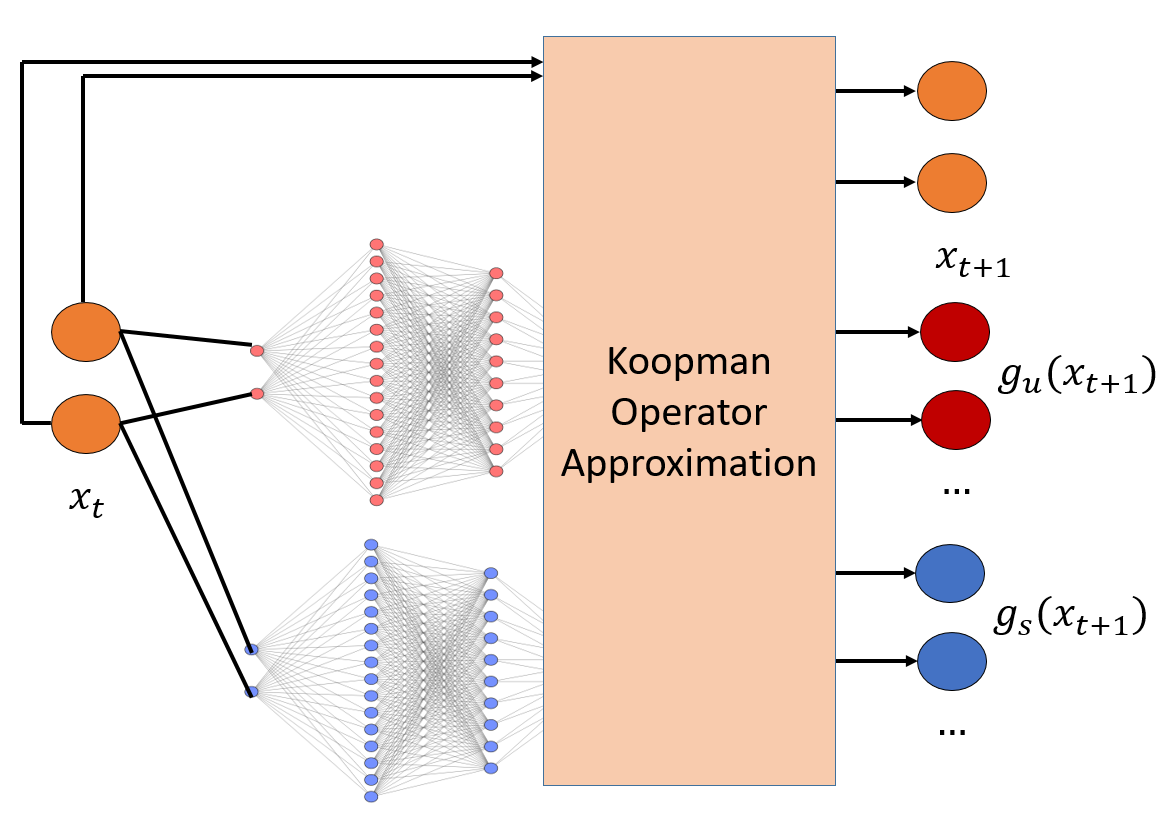}
    \caption{Resultant model architecture of the Joint Model described in Algorithm \ref{algo:ml}.}
    \label{fig:hybrid_arch}
\end{figure}

The observable functions produced by these networks are then concatenated as shown in Fig. \ref{fig:hybrid_arch}. This new linear transition matrix can be recomputed utilizing the Direct Encoding method. Let $g_i(x;\theta_i)$ be the $i$-th observable in the form of a neural network with parameters $\theta_i$. The inner product comprising the matrix $R$ is give by

\begin{equation}
    R=\{R_{ij}\}=\lbrace\langle{g_i}(x;\theta_i),g_j(x;\theta_j)\rangle\rbrace
\end{equation}
Similarly, 
\begin{equation}
    Q=\{Q_{ij}\}=\lbrace\langle{g_i}(x;\theta_i) \circ f(x),g_j(x;\theta_j)\rangle\rbrace
\end{equation}


Note that simulation data are used only for finding effective observables and not for obtaining the state transition $A$ matrix. The $A$ matrix is obtained from DE by using the neural net observables as shown in the above equation. Algorithm \ref{algo:ml} summarizes this procedure,  referred to in latter sections as the SSOG Model, as it joins together two neural network models that are constructed to generate observables specific to the subspaces they are trained on.


\begin{algorithm} 
    \caption{Training scheme for the prediction models. Notation is explained in Table \ref{tab:vardefinitions}.}
    \label{algo:ml}
    \KwIn{}
    $G_u$: $x_u \sim X_u$; $G_s$: $x_s \sim X_s$; $G_{joint}$: $x \sim X$; \\ 
    $F_u$: $z_{u_k}$; $F_s$ : $z_{s_k}$; $F_j$: $z_{j_k}$\\
    Parameters: (weight variables);\\
    \KwOut{}
    $G_u$: $g_u$; $G_s$: $g_s$; $G_{joint}$: $g_{joint}$\\
    $F_u$: $\hat{z}_{u_{k+1}}$; $F_s$: $\hat{z}_{s_{k+1}}$; $F_j$: $\hat{z}_{j_{k+1}}$ \\
    \For{number of epochs}{
    Sample $x_*$ from $X_*$ \\
    Generate observable function vector $g_*$, by inputting $x_*$ through nonlinear layers \\
    Append state vector to observable vector $z_* = x_* \oplus g_*$ \\
    Estimate augmented output through linear activation $F_*(z_{*_k})$ \\
    Forward pass through MSE Loss function $L(z,\hat{z})$; \\
    Backward pass: update parameters for observable functions ($W_u, W_s$); \\
    }
    \textbf{then}{ \\
    Calculate augmented input $g_{u_k}$ and $g_{s_k}$ for $x_{k_{DE}} ~ X_{DE}$ for all $X_{DE}$ \\
    Append augmented inputs $z_{k_{DE}}$ = $x_{k_{DE}} \oplus g_{u_k} \oplus g_{s_k}$\\
    Calculate augmented output $g_{u_{k+1}}$ and $g_{s_{k+1}}$ for $x_{{k+1}_{DE}} ~ X_{DE}$ for all $X_{DE}$ \\
    Append augmented outputs $z_{{k+1}_{DE}}$ = $x_{{k+1}_{DE}} \oplus g_{u_{k+1}} \oplus g_{s_{k+1}}$\\
    Calculate state transition matrix for linear layer of joint model
    }
\end{algorithm}

\begin{table} 
    \small
    \centering
    \begin{tabular}{c|c}
    \hline
        Notation & Definition \\
        \hline
         x & state vector \\
         $x_u$ & state vector belonging to unstable region \\
         $x_s$ & state vector belonging to stable region\\
         $G_*$ & Observable function model trained on the $*$ dataset \\
         $F_*$ & Linear regression model trained on $*$ dataset \\
         $x_{*_k}$ & state vector at initial time step \\
         $x_{*_{k+1}}$ & state vector at next time step \\
         $W_*$ & weights for a model ($G_*$ and $F_*$) \\
         $z$ & lifted state vector \\ 
         $L$ & Loss function \\
         $*$ & Subscript denotes being used \\&for stable ($s$), unstable ($u$) regions \\
    \end{tabular}
    \caption{Notation and definitions for variables indicated in different parts of this paper.}
    \label{tab:vardefinitions}
\end{table}

\section{Experiments}
\label{results}
 
The computer creating these models has a AMD Ryzen 7 3700X 8-Core Processor 3.60 GHz, and a NVIDIA GeForce GTX 1060 3GB Graphics Card. The models are created using PyTorch. The SSOG models have two hidden layers with ReLU activation units, and the final layer utilizes a linear layer, which represents the Koopman Operator's linear transition matrix. The width of the first hidden layer is 16 units, and the width of the second hidden layer is 10 units. These hidden layers utilize rectified linear activation units. In the case of 40 observables, the SSOG model has 20 observables for both stable and unstable regions. In the case of 80 observables, the division is done similarly. In addition to these observable functions, the original state variables are included as part of the model as well. The loss function for the neural networks is a standard mean squared error loss function of 
\begin{equation}
    L_{MSE} =\frac{1}{N} \sum_t | z_{t+1} - A z_t |^2
\end{equation}
where $z$ is the lifted state variable, and $A$ is the weights of the final linear layer. The model is trained using an Adam optimizer with a learning rate of $\alpha = 0.01$. Hyperparameters were equivalent between models.

The SSOG model with DE is compared to the SSOG model without the use of DE, where the weights of the linear activation layer is computed from a least squares regression. It is also compared to an Aggregate model which utilizes the same neural network structure but aggregates the dataset together, training a single network that produces twice the number of observable functions and is not subspace specific. A DE version of the Aggregate model is also formulated, recomputing the final layer of the model using DE.

Standard EDMD models are constructed to be of equivalent order to the SSOG and Aggregate without deep learning. The EDMD model's observables are radial basis functions that are uniformly distributed for individual states between minimum and maximum value for that state in the dataset. These radial basis functions are of the form
 \begin{equation}
    \psi = e^{\omega (x_i - x_{i_a})^2} 
    \label{eq: RBF}
\end{equation}
where $\omega$ is a parameter that was set to equal 1 for all state variables and $x_i$ is the $i$th state variable, and $x_{i_a}$ is the center of the radial basis function.
 
\begin{figure}
    \centering
    \includegraphics[width = 0.8\linewidth]{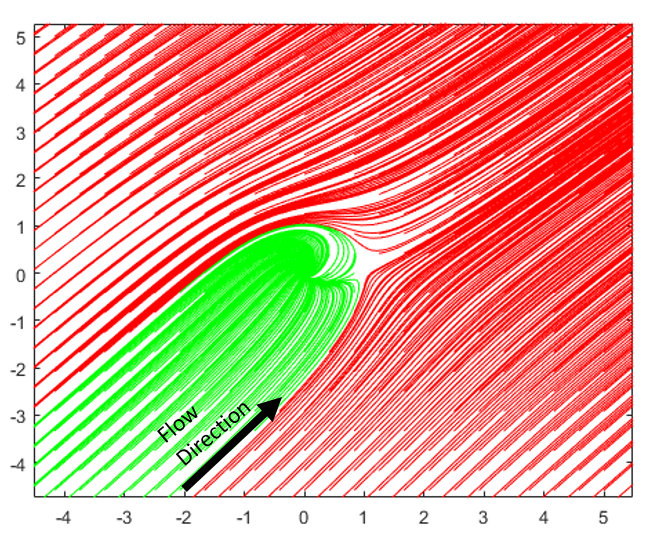}
    \caption{Phase Plot visualization of initial conditions that yield stable versus unstable trajectories. The green trajectories symbolize the stable region, and the red trajectories symbolize the unstable region.}
    \label{fig:2ord_pp}
\end{figure}

We introduce a second order nonlinear system:
\begin{align*}
    \dot{x} &= -x + x^2 + y^2 \\
    \dot{y} &= -y + y^2 + x^2 - x \\
\end{align*}
This system has effectively two regions, a stable region and an unstable region, visualized with a collection of trajectories in Fig. \ref{fig:2ord_pp}.

 \subsection{Prediction Error}

The prediction error for this system is calculated separately for stable and unstable region time series trajectories with a set of test data consisting of initial conditions not included for either DE nor the aggregate trajectory dataset. The equation used for prediction error is sum of squared error for the state variables, not including observables of the system. That is
\begin{equation}
    E_{sse} = \sum_{i=0}^N (x_i - \hat{x}_i)^2
\end{equation}
where $i$ is the $i$th state variable for a given time step. The estimated state variable, $\hat{x}$, is the prediction from the model, where the ground truth is denoted as $x$. This prediction error is not of the total lifted state, but of the state variables belonging to the original nonlinear system which are shared between all models.

\begin{figure} 
\includegraphics[width = \linewidth]{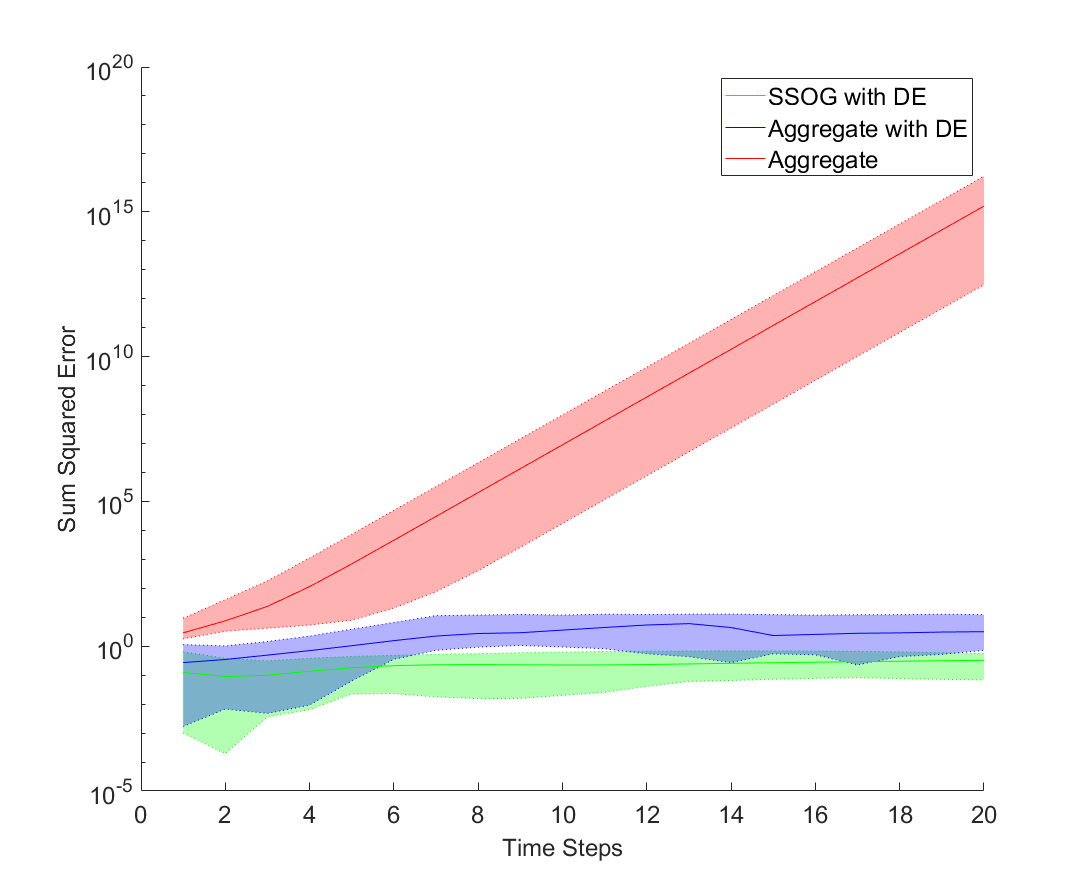}
    
\caption{Prediction error for 100 trajectories beginning at initial conditions of a test set within the stable subspace. The shaded region represents the variation between minimum and maximum SSE at the given time step for the listed model. The solid lines represent the average sum of squared error. Comparison is between the SSOG with Direct Encoding, and Aggregate Model with and without DE. The models plotted utilize 40 observables in addition to the two state variables.}

\label{fig:sse2}
\end{figure}

The prediction error is visualized in the set of figures, Fig. \ref{fig:sse2}. The two figures compare the SSOG with DE and other learned models. Further comparisons using this test set are outlined in Table \ref{tab:prederror}. The best performing result of the labeled order for each subspace for each set of models is in bold.

\begin{table} [!ht]
\caption{Average SSE prediction error after 1 and 10 time steps for trajectories in set of test data.}
\label{tab:prederror} \begin{center} \begin{tabular}{c c c }
\hline \hline 

Method & Stable & Unstable  \\
\hline 

\multicolumn{3}{c}{\textit{40 Observables, 1 Time Step}} \\
SSOG & 2.84 & 6.15\\
SSOG with DE & \textbf{0.12} & \textbf{0.26}\\
Aggregate & 2.90 & 5.23 \\
Aggregate with DE & 0.27 & 1.44\\
EDMD & $3.79 \times 10^{3}$ & $2.06 \times 10^{3}$ \\
\hline 
\multicolumn{3}{c}{\textit{80 Observables, 1 Time Step}} \\
SSOG& 4.16 & $4.80 \times 10^{5}$ \\
SSOG with DE & \textbf{0.11} & \textbf{0.36}\\
Aggregate & 2.20 & 2.83 \\
Aggregate with DE & 0.60 & 1.12\\
EDMD & $1.28 \times 10^7$ & $7.71 \times 10^{6}$ \\
\hline
\multicolumn{3}{c}{\textit{40 Observables, 10 Time Steps}} \\
SSOG & $2.50 \times 10^{36}$ & $6.8 \times 10^{131}$  \\
SSOG with DE & \textbf{0.22} & $6.8 \times 10^{131}$\\
Aggregate & 9.00 $\times 10^6$ & $6.8 \times 10^{131}$ \\
Aggregate with DE & 3.60 & $6.8 \times 10^{131}$\\
EDMD & $2.15 \times 10^{35}$ & $6.8 \times 10^{131}$ \\
\hline 
\multicolumn{3}{c}{\textit{80 Observables, 10 Time Steps}} \\
SSOG & $5.71 \times 10^{10}$ & $6.8 \times 10^{131}$ \\
SSOG with DE & \textbf{0.18} & $6.8 \times 10^{131}$\\
Aggregate & $1.06 \times 10^8$ & $6.8 \times 10^{131}$ \\
Aggregate with DE & 11.63 & $6.8 \times 10^{131}$\\
EDMD & $2.10 \times 10^{73}$ & $6.8 \times 10^{131}$ \\
\hline
\end{tabular} \end{center} \end{table}

\subsection{Subspace Boundary Formation}

Finding the boundary between stable and unstable regions is important for analyzing nonlinear dynamics. Here we compare each modeling method in terms of the accuracy in finding the boundary. Consider the following quotient:

\begin{equation}
\xi = \frac{||z_u||}{||z||} 
\label{eq:instabquotient}
\end{equation}
where
\begin{equation}
z_u = W_u^T z 
\end{equation}
If $z$ is in a stable region and the matrix $W_u$ spanning the unstable region is accurate, then the quotient $\xi$ must be 0. On the other hand, if $W_u$ is inaccurate, then some fraction of the component will enter the stable region with non-zero $\xi$. Fig \ref{fig:2ord_pp} shows the plots of this quotient indicating how accurately the true boundary is recreated.

\begin{figure*} [!ht]
    \centering
    \includegraphics[width=0.77\textwidth]{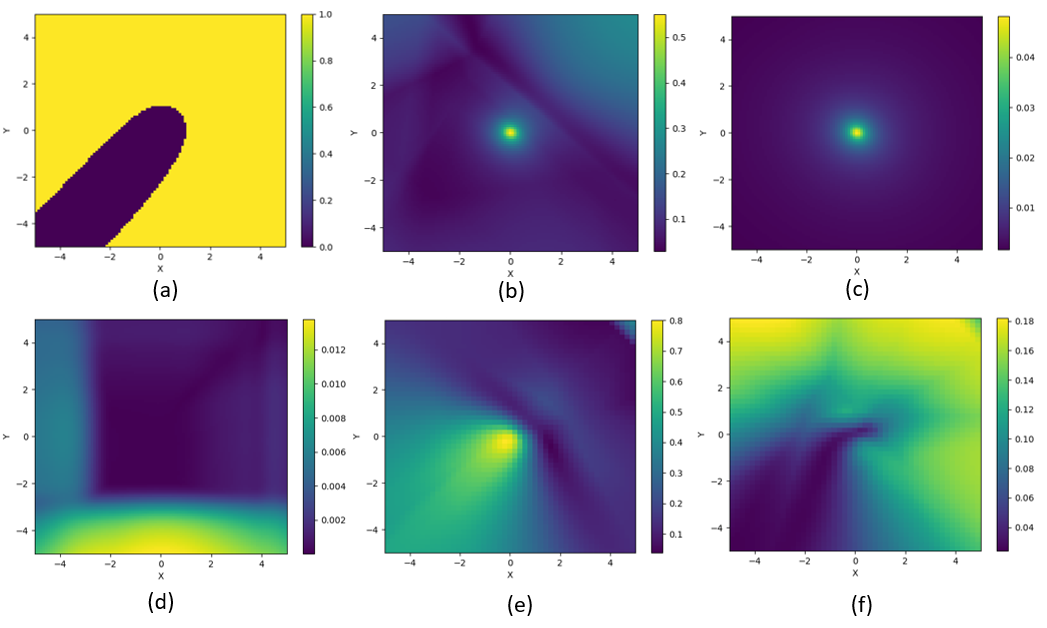}
    \caption{Instability quotient plots for models of the system. Top: (a) Ground Truth assuming complete separation, (b) Aggregate Model, (c) Aggregate Model with Direct Encoding. Bottom: (d) EDMD, (e) SSOG Model, (f) SSOG Model with Direct Encoding. }
    \label{fig:iqarr}
\end{figure*}

\section{Discussion}
\label{discussion}

The first key result found from Table \ref{tab:prederror} is that using SSOG in combination with DE yields the highest accuracy is obtained demonstrating the usefulness of subspace specific observable functions. Secondly, the DE method is shown to drastically lower the prediction error for both Aggregate and SSOG models by several orders of magnitude. This result matches the expectation that DE removes biases from over- or undersampling of a region given its formulation using inner products over a domain. Furthermore, the DE method is demonstrated to have a significant impact even with increases in order of the system. Notably however, none of the models are capable of predicting trajectories in the unstable subspace.

When observing Table \ref{tab:prederror}, it is notable that the prediction error does not necessarily improve with increases in order of the system. This can be explained as the tabulated results are prediction errors of the state variables, which are shared across all models. As the order of the systems increase, a least squares estimation would weight the prediction error of the state variables similarly to the observable functions used. The additional observables may detract from the desire to accurately predict the state variables for a significant time horizon. 

Because the prediction models used are finite order approximations of the Koopman operator, we expect inaccuracies to arise when using these models to locate boundaries between regions. For that reason, the instability quotient is introduced. Assuming complete separation of dynamic modes, the ratio of unstable projection to state vector (instability quotient) should be 0 for the stable region, and 1 for the unstable region, as shown in plot (a) of Fig. \ref{fig:iqarr}. However, due to the approximation of the system as a linear system, this discrete switch becomes blurred, depending on the accuracy of the model which is shown in the other plots. The SSOG Model with DE demonstrates that the instability quotient increases significantly in the unstable region. This does not occur for the other models. This indicates that the observables learned for the SSOG Model are effective for their specific subspaces. The DE method is also key in this result as it demonstrates a drastic change in the dynamic model.



\section{Conclusion}
\label{conclusion}

In this paper, we presented the subspace specific observable generation (SSOG) method for learning an efficient set of observable functions to lift the state space of nonlinear dynamic systems. In combination with an existing method, Direct Encoding (DE), SSOG was shown to improve accuracy given certain conditions. Separately, SSOG with DE demonstrates a capacity to be used as an analysis tool for finding borders between subspaces based on analytical foundations. The work has a clear direction for improvement; the current network structure and training does not enable the subspaces to be fully separated as the observable functions learned are not zero outside of their respective regions. This would be a notable direction to explore in the future to further improve results.

\bibliographystyle{IEEEtran}
\bibliography{IEEEabrv, main.bib}
\end{document}